\documentclass[review]{elsarticle}

\usepackage{multirow}
\usepackage{array,balance}
\usepackage{graphicx}
\usepackage{amsmath}
\usepackage{amssymb}
\usepackage{xcolor}

\journal{Neurocomputing}

\begin{document}

\begin{frontmatter}

\title{Towards More Effective PRM-based Crowd Counting via A Multi-resolution Fusion and Attention Network}

\author[mymainaddress]{Usman Sajid}

\author[mysecondaryaddress]{Guanghui Wang\corref{mycorrespondingauthor}}
\cortext[mycorrespondingauthor]{Corresponding author}
\ead{usajid@ku.edu; wangcs@ryerson.ca}

\address[mymainaddress]{Department of Electrical Engineering and Computer Science, University of Kansas, Lawrence, KS, USA, 66045}
\address[mysecondaryaddress]{Department of Computer Science, Ryerson University, Toronto, ON, Canada M5B 2K3}

\begin{abstract}
The paper focuses on improving the recent plug-and-play patch rescaling module (PRM) based approaches for crowd counting. In order to make full use of the PRM potential and obtain more reliable and accurate results for challenging images with crowd-variation, large perspective, extreme occlusions, and cluttered background regions, we propose a new PRM based multi-resolution and multi-task crowd counting network by exploiting the PRM module with more effectiveness and potency. The proposed model consists of three deep-layered branches with each branch generating feature maps of different resolutions. These branches perform a feature-level fusion across each other to build the vital collective knowledge to be used for the final crowd estimate. Additionally, early-stage feature maps undergo visual attention to strengthen the later-stage channel's understanding of the foreground regions. The integration of these deep branches with the PRM module and the early-attended blocks proves to be more effective than the original PRM based schemes through extensive numerical and visual evaluations on four benchmark datasets. The proposed approach yields a significant improvement by a margin of $12.6\%$ in terms of the RMSE evaluation criterion. It also outperforms state-of-the-art methods in cross-dataset evaluations.
\end{abstract}


\end{frontmatter}


\section{Introduction}
\begin{figure}[t]
	\begin{minipage}[b]{0.490\columnwidth}
		\begin{center}
			\centerline{\includegraphics[width=0.795\columnwidth]{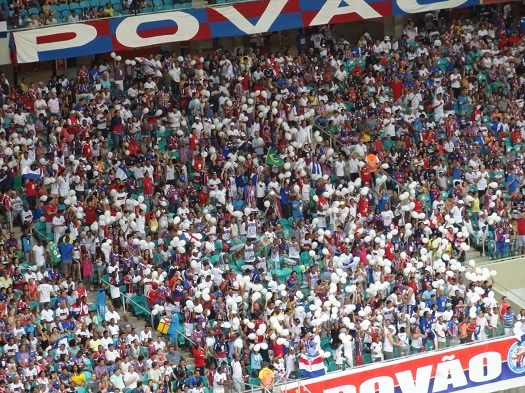}}
		\end{center}
	\end{minipage}
	\begin{minipage}[b]{0.490\columnwidth}
		\begin{center}
			\centerline{\includegraphics[width=0.795\columnwidth]{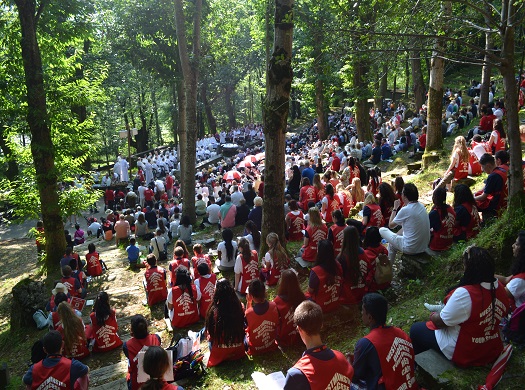}}
		\end{center}
	\end{minipage}
		\begin{minipage}[l]{0.490\columnwidth}
		\begin{center}
			\centerline{\includegraphics[width=0.795\columnwidth]{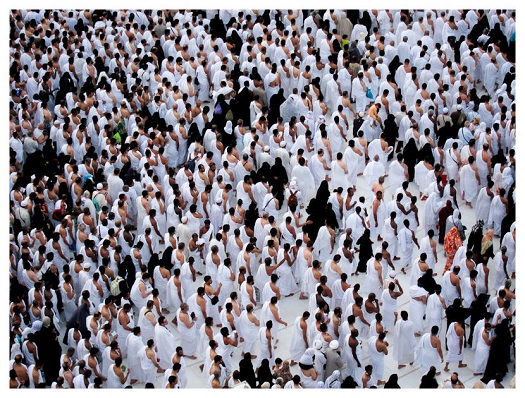}}
		\end{center}
	\end{minipage}
	\begin{minipage}[r]{0.490\columnwidth}
		\begin{center}
			\centerline{\includegraphics[width=0.795\columnwidth]{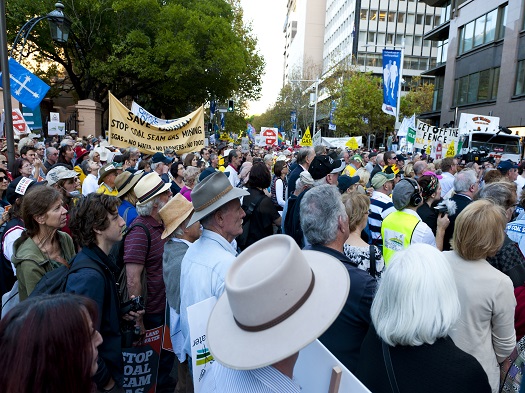}}
		\end{center}
	\end{minipage}			
	\caption{\footnotesize{Sample crowd images. Estimating the crowd in these images comes up with many major challenges and problems including huge fluctuation of the crowd-density in and across different images, different image resolution, far-reaching wide perspective, and severe occlusions.
	}}
	\label{fig:fig1}
\end{figure}

Deep learning has achieved exceptional success in many computer vision applications like classification and detection \cite{cen2021deep,li2021sgnet,ma2020location}. Crowd counting aims to estimate the total number of people in a given static image. This is a very challenging problem in practice since there exists a significant difference in the crowd number in and across different images, varying images resolution, large perspective, and severe occlusions \cite{sajid2021audio}, as shown in Fig. \ref{fig:fig1}. Accurate crowd counting can help to effectively organize large crowd gatherings. Convolutional Neural Networks (CNNs) have proven to be very effective in dealing with these issues. The state-of-the-art crowd counting techniques are based on either counting-by-regression or the density-map estimation. 

Counting-by-regression \cite{wang2015deep,sajid2020zoomcount,sajid2020plug} schemes learn the mapping of the input image or patch to its crowd count, whereas the density-map estimation methods \cite{zhang2016single,cascadedmtl,sam2017switching,ranjan2018iterative,sajid2020multi,wan2019residual,xu2019learn} yield the crowd-density value per input image pixel that are summed to get the image final crowd count. In general, counting-by-regression schemes do not perform reasonably well without any special and additive mechanism. On the other hand, density-map based methods rely heavily on the accurate density-map generation for the training images from the available ground-truth dot-map annotations for the learning process of their models. In principle, a point-spread function (e.g. Multivariate Gaussian Kernel) is being deployed to produce pixel-wise crowd-density values. Although the density-based methods produce reasonable results, the amount of Gaussian spread challenge limits their overall performance and relatively compromises their efficacy. Some works \cite{shami2018people,li2019headnet} also aim to detect people by their head or body using some well-established CNN-based detectors for the crowd counting purpose, however, few pixels per head in medium to high-dense crowd images make it almost impractical to achieve.

\begin{figure*}
	\begin{minipage}[b]{1.0\textwidth}
		\begin{center}
			\centerline{\includegraphics[width=1.0\textwidth]{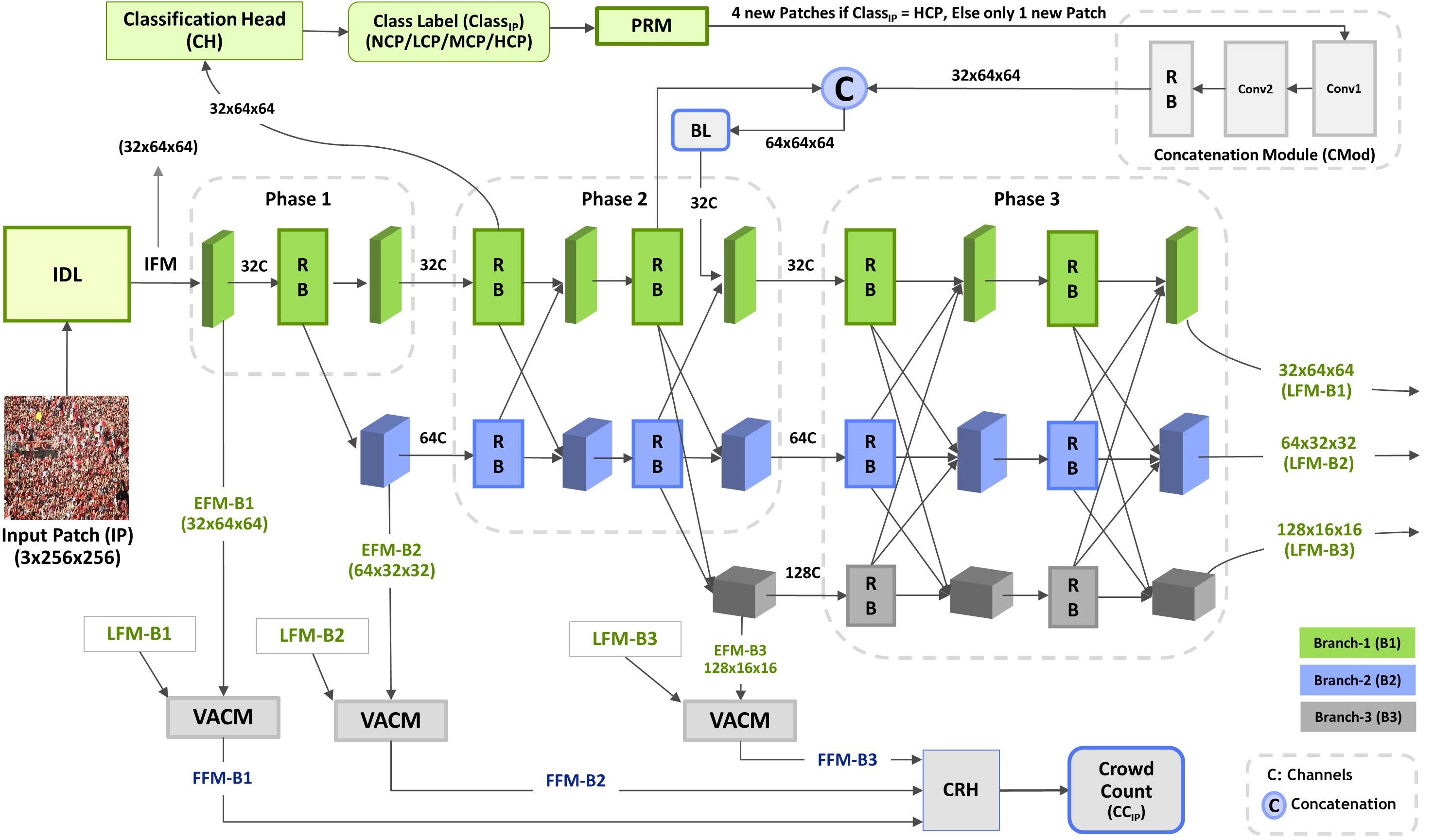}}
		\end{center}
	\end{minipage}		
	\caption{The proposed multi-task feature-level fusion based crowd counting network. The $256\times256$ input patch ($IP$) passes through the initial deep layers ($IDL$) to output the initial feature maps ($IFM$) that are fed into the phase-based multi-branch network. Each of the three branches/columns maintains its original resolution throughout that branch \cite{sun2019deep,wang2020deep}. The output channels from the first residual block (RB) of the (Branch-1, Phase-2) are branched out to make the classification head ($CH$). The $CH$ classifies the patch ($IP$) 4-way according to its crowd-density level. This labeling process is being used by the PRM module \cite{sajid2020plug} to decide if appropriate rescaling and generation of new patches are required or not. The resultant patch(es) then passes through the concatenation module ($CMod$) to generate the channels ($C$) that are concatenated back into the network after adjusting the number of channels via the Bottleneck layer ($BL$). The network branches also do the feature-level fusion regularly to form the model collective knowledge. After the remaining processing, the Phase-3 outputs three later-stage feature maps (LFM-B1, LFM-B2, LFM-B3). Each LFM passes through the VACM module along with their respective branch early-stage feature maps (EFM-B1, EFM-B2, EFM-B3). The VACM generated final feature maps (FFMs) are eventually inserted into the concatenation-based Crowd Regression Head ($CRH$) to obtain the final crowd count (${CC}_{IP}$) for the input Patch ($IP$).} 
   
    \label{fig:fig2}
\end{figure*}

Recently, Sajid \textit{et al.} \cite{sajid2020plug} proposed a counting-by-regression based method that uses a lightweight Patch Rescaling Module (PRM) to rescale the input image or patch accordingly based on its crowd-density level before the crowd estimation. They also proposed PRM-based schemes, which performed reasonably better as compared to other state-of-the-art methods. Although these schemes produce state-of-the-art results, they carry the following key shortcomings:

\begin{itemize}
  \item They maintain a single branch/column architecture with only-one scale focus that limits their achievable performance and potential effectiveness. 
  \item These models only utilize the PRM based input-level multi-resolution rescaling process, while lacking the beneficial feature-level multi-resolution process.
  \\
\end{itemize}

In this work, we present a new multi-resolution feature-level fusion based multi-task and visually-attended crowd counting method that aims to address the major crowd counting challenges as well as further explores and pushes the boundaries of the PRM module by tackling its shortcomings as mentioned above. The proposed PRM based scheme, as shown in Fig. \ref{fig:fig2}, incorporates multiple columns or branches, each with feature maps of different resolutions. Based on the high-resolution networks \cite{sun2019deep,wang2020deep}, these branches perform the fusion or sharing across each other on a regular basis to form a collective knowledge that improves the overall network performance. We also perform the visual-attention process on the early-stage feature maps from each of these branches to boost later-stage channel's understanding of foreground and background. On the other hand, the PRM based rescaling of the input image or patch has been used to select the appropriate input-level scale based on its crowd-density level. The experiments (Sec. \ref{expi}) demonstrate the proposed PRM utilization to be remarkably more effective as compared to its previous implementations \cite{sajid2020plug,sajid2020zoomcount}. The key contributions of this paper are listed as follows:

\begin{itemize}\setlength\itemsep{0.3em}
  \item We propose a new multi-resolution and multi-task PRM based visually-attended crowd counting network that effectively addresses major crowd counting challenges, including the issues of crowd-like background regions and huge crowd-variation.
  \item We deploy the plug-and-play PRM module so as to further push its boundaries and utilize it more effectively as compared to its previous deployments.
  \item We employ the visual attention mechanism in a unique and effective way on early-stage feature maps that facilitates the later-stage channels to better understand the foreground regions.
  \item The proposed scheme outperforms the current-best methods in most cases, including the original PRM based schemes as demonstrated via both numerical and visual experiments on four benchmark datasets. The proposed model shows an improvement of up to $12.6\%$ in terms of RMSE evaluation while performing almost equally best for the MAE evaluation criterion. During the cross-dataset evaluation, the proposed method outperforms the state-of-the-art.
\\  
\end{itemize}

\section{Related Work}
Crowd counting has many different key challenges including image resolution and crowd-density fluctuation in and across different images, extreme occlusions, far-reaching crowd perspective within an image, cluttered no-crowd regions in the images. Many classical schemes were proposed that generally aimed to solve this task as either a regression or a detection problem. Regression-based counting \cite{chan2009bayesian,chen2012feature,ryan2009crowd} learns the local crowd features mapping function for the given image to regress the crowd count. But they fail to reasonably comprehend the key crowd counting challenges. Detection-dependent \cite{ge2009marked,li2008estimating,wang2011automatic,wu2005detection} schemes detect hand-crafted features to estimate the crowd count. These methods become impractical and ineffective in the case of high-dense crowd images because of the very small area per head or person.

Recently, Convolutional Neural Networks (CNNs) based approaches are the backbone of almost all crowd counting methods. Generally, we can divide them into three classes: Detection-dependent, Regression-dependent, and density-map estimation \cite{sajid2020zoomcount}. Well established object detectors (like YOLO et al. \cite{redmon2016you,ma2020mdfn}) are being employed by the detection-dependent \cite{shami2018people,li2019headnet} methods to identify the people in the given image, and then sum all the detections to give the final crowd count. However, they fail for the images with high-density crowd due to smaller area per person or head. Shami \textit{et al.} \cite{shami2018people} performed the weighted average of the CNN-based detections to produce the final people count. Li \textit{et al.} \cite{li2019headnet} employed the CNN based adaptive head detection method that used the context-based information for the crowd estimation. 

Regression-dependent \cite{wang2015deep,sajid2020zoomcount,sajid2020plug} methods map the input image directly to its crowd estimate. These methods also do not generalize well without any specialized and supportive mechanism. Wang \textit{et. al} \cite{wang2015deep} used the AlexNet \cite{krizhevsky2012imagenet} based scheme to directly regress and estimate the people number. Density-map estimation \cite{zhang2016single,cascadedmtl,sam2017switching,ranjan2018iterative,wan2019residual,xu2019learn} methods estimate the crowd-density value per pixel. The pixel-wise density values get added to produce the image's final crowd number. These methods first require the conversion of the ground-truth crowd dot-maps to their respective density-maps using a point-spread function (e.g. multivariate Gaussian distribution kernel) for the model training. However, the spread or kernel size remains a major issue that is vital for their training process quality. MCNN \cite{zhang2016single}, a density-map estimation method, uses three columns or branches, each with different filter sizes to account for the respective crowd-scale. These branches are merged together at the end to generate the final crowd density-map. 

Sam \textit{et al.} \cite{sam2017switching} designed the Switch-CNN, another density-map method, that contains a (CNN-based) switch classifier that labels each input image or patch and routes it accordingly to one of three available specialized crowd regressors. Sindagi \textit{et al.} \cite{cascadedmtl} devised a cascaded network that computes the crowd by first classifying the input 10-way based on its crowd-density level. Ranjan \textit{et al.} \cite{ranjan2018iterative} designed a 2-branch network where the lower-resolution branch or column helps the higher-resolution for better crowd calculation. Li \textit{et al.} \cite{li2018csrnet} presented the CSRNet to obtain the context-based information through the dilation based deep layers for effective crowd estimation. Idrees \textit{et al.} \cite{idrees2018composition} designed the density-map estimation network that performs better for the crowd counting task by simultaneously solving the crowd localization as well as for the crowd-density estimation. Xu \textit{et al.} \cite{xu2019learn} designed the density-map based method that used the patch-based density-maps, and grouped them into several levels based on the crowd-density, followed by the online learning mechanism for the automatic normalization process with a multi-polar center loss.  Recently, Sajid \textit{et al.} \cite{sajid2020zoomcount,sajid2020plug} proposed a simple plug-and-play patch rescaling module (PRM) that rescales the input image or patch based on the respective crowd-density label.

Although recent works have produced very promising results, none of them make full use of their capabilities nor generalize well to hugely varying crowd-density and other major challenges. In this work, we focus on addressing the above crowd estimation challenges and issues.

\begin{table}[htbp]\small
	\begin{center}
	\begin{tabular}{|c|c|c|}
    \hline
 Layer & Output & Filters (F) \\ \hline \hline
\multicolumn{3}{|c|}{Initial Deep Layers (IDL)} \\ \hline
 $IP$ & $3 \times 256 \times 256$ &     \\ \hline
 & $64 \times 128 \times 128$ &  ($3 \times 3$) conv, s = 2, p = 1, $64$F    \\ \hline
  & $64 \times 64 \times 64$ &  ($3 \times 3$) conv, s = 2, p = 1, $64$F    \\ \hline
 $IFM$ & $32 \times 64 \times 64$ &  ($1 \times 1$) conv, s = 1, p = 0, $32$F    \\ \hline \hline
 \multicolumn{3}{|c|}{Classification Head (CH)} \\ \hline
  & $32 \times 64 \times 64$ &     \\ \hline
  & $64 \times 32 \times 32$ &  ($3 \times 3$) conv, s = 2, p = 1, $64$F    \\ \hline
  & $32 \times 16 \times 16$ &  ($3 \times 3$) conv, s = 2, p = 1, $32$F    \\ \hline
  & $32 \times 8 \times 8$ &  ($2 \times 2$) Global Avg. Pooling, s = 2 \\ \hline
  & 1024D FC &  -    \\ \hline
  & 4D FC, Softmax &  -    \\ \hline \hline
\multicolumn{3}{|c|}{Concatenation Module (CMod)} \\ \hline
  & $3 \times 256 \times 256$ &     \\ \hline
 Conv1 & $64 \times 128 \times 128$ &  ($3 \times 3$) conv, s = 2, p = 1, $64$F    \\ \hline
 Conv2 & $32 \times 64 \times 64$ &  ($3 \times 3$) conv, s = 2, p = 1, $32$F    \\ \hline
 RB & $32 \times 64 \times 64$ &  -    \\ \hline \hline
\multicolumn{3}{|c|}{Crowd Regression Head (CRH) (Continued from Fig. 5(b))} \\ \hline
& $64 \times 32 \times 32$ &     \\ \hline
 & $64 \times 16 \times 16$ &  ($3 \times 3$) conv, s = 2, p = 1, $64$F    \\ \hline
& $64 \times 8 \times 8$ &  ($2 \times 2$) Avg. Pooling, s = 2    \\ \hline
& 1024D, FC &  -    \\ \hline
& 1D, FC  &  -    \\ \hline
	\end{tabular}
	\end{center}
\caption{\footnotesize Configurations of IDL, CH, CMod, and CRH modules. Each \textit{conv} operation denotes the Convolution-BN-ReLU sequence \cite{huang2017densely}. (s: stride, p: padding, FC: Fully Connected)}

\label{table:table1}
\end{table}

\section{Proposed Approach}
This work aims to address the major crowd counting challenges (e.g., huge difference in image and scale resolution, severe occlusions, far-reaching perspective changes, etc.) as well as to exploit the PRM module more effectively while mitigating its shortcomings. The proposed multi-task network, as shown in Fig. \ref{fig:fig2}, contains three deep-layered branches with different resolutions feature maps (channels). The feature-level fusion \cite{sun2019deep,wang2020deep} occurs in between these branches on a regular basis, which helps them to form a collective knowledge about the input image or patch. As their final output, each of these branches produces early- (EFM) and later-stage (LFM) feature maps. The EFMs undergo visual attention process before concatenation with the LFM channels via the visual attention and concatenation (VACM) module. The visually-attended EFMs enable the LFMs to clearly distinguish between foreground and background image regions. The concatenated final feature maps (FFMs) then pass through the crowd regression head (CRH) to output the final crowd estimate. The PRM module has been deployed to rescale the input to its appropriate scale. To start with, we first divide the input image into fixed-size $256 \times 256$ non-overlapping patches. Following that, we estimate the crowd count for each resultant patch separately using the proposed scheme. Finally, the image total people count is equal to the summation of all its patches crowd estimates. 

The input patch ($IP$) first proceeds through the initial deep layers (IDL), as detailed in Table \ref{table:table1}, to produce the initial feature maps (IFMs). The IDL also reduces the input resolution  to $1/4$ (from $256 \times 256$ to $64 \times 64$). Next, these feature maps are routed to Phase-1 of the multi-branch network. The phase-wise multi-branch network comprises of five key components: Multi-resolution branches and phases, Multi-resolution fusion, PRM module deployment, VACM module, and Crowd Regression Head (CRH). They are detailed as follows.

\subsection{Multi-resolution Branches and Phases}
The network comprises of three phases (Phase-1, 2, and 3). The initially generated feature maps (IFMs) or channels pass through the phase-wise organized multi-branch deep layers, starting from Phase-1. Each phase contains the total number of multi-resolution branches equal to its phase number. Consequently, Phase-1, 2, and 3 contain one, two, and three deep branches, respectively. Each branch also maintains its channels resolution throughout that branch \cite{sun2019deep,wang2020deep}. These branches also perform feature-level fusion across each other on a regular basis to form a collective knowledge-based learning process, as detailed in the next subsection \ref{fusion}. The channel resolution and the total number of channels in a specific branch depend on the highest-resolution branch configuration (Branch-1). The channel resolution decreases by half in each subsequent lower-resolution branch. However, the total number of channels increases $2 \times$ times as we move from higher to lower-resolution branches. Thus, Branch-1, 2, 3 contain ($32 \times 64 \times 64$), ($64 \times 32 \times 32$), ($128 \times 16 \times 16$) channels respectively, where it is denoted by ($Channels \times Width \times Height$).

\begin{figure}[t]
	\begin{minipage}[b]{1.0\columnwidth}
		\begin{center}
			\centerline{\includegraphics[width=0.8\columnwidth]{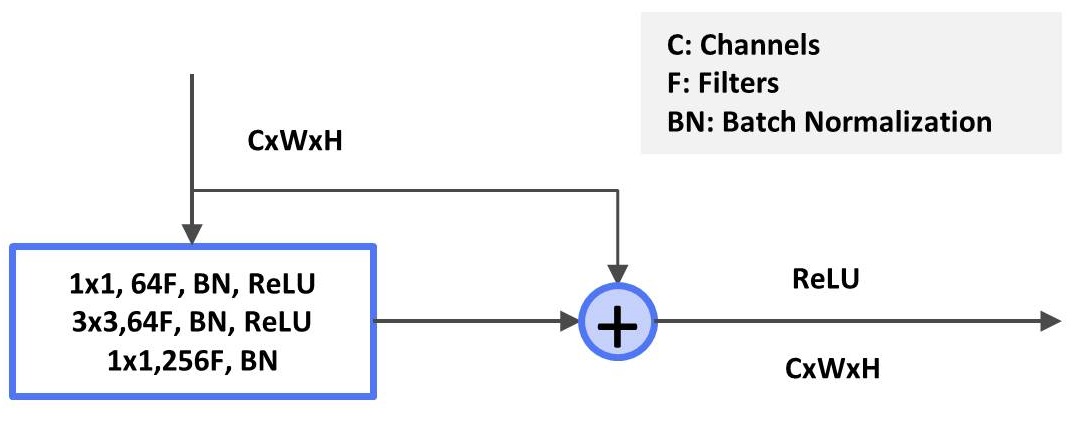}}
		\end{center}
	\end{minipage}
	\caption{\footnotesize{3-Layered Residual Unit \cite{he2016deep} being used in the Residual Block (RB). Each RB is composed of four such units.
	}}
    \label{fig:fig3}
\end{figure}

Each phase also contains the Residual Blocks (RB). Each RB comprises of four residual units \cite{he2016deep} that are 3-layer based residual modules as shown in Fig. \ref{fig:fig3}. Batch Normalization (BN) \cite{ioffe2015batch} and the non-linear ReLU activation \cite{nair2010rectified} follow each convolution operation. The number of such RB modules remains the same in each branch of a specific phase.

Eventually, the phase-based three-branch structure outputs six separate blocks of feature maps; three early- (EFM) and three later-stage (LFM) feature maps. The phase-3 output blocks from the respective branch Bx (x=1,2, or 3) serve as the LFM-Bx channels. To obtain the early-stage EFM-Bx channels, we take the very first channels being produced in that specific branch Bx. These channels proceed forward for further processing.

\subsection{Repeating Multi-resolution Fusion}
\label{fusion}
The branches in a phase regularly share their channels across each other via the summation-based fusion. This sharing process helps in learning and building the collective information and knowledge from all branches that naturally enhances the generalization potential of the proposed network towards huge crowd-density and scale variation. The higher-branch channels are fused into the lower-resolution channels using the $(3 \times 3, stride-2, padding-1)$ convolution(s) to down-size the resolution accordingly \cite{sun2019deep,wang2020deep}. To fuse the Branch-1 channels into the Branch-2, Branch-1 channels are down-sampled by using one such $3 \times 3$ convolution operation. Similarly, the Branch-1 fusion into Branch-3 requires this convolution operation twice. To fuse the lower-resolution channels into the higher-level branch, bilinear upsampling has been applied to lower-resolution features to up-size them accordingly before the fusion process \cite{sun2019deep,wang2020deep}.

\subsection{PRM Module Deployment}
The purpose of the Patch Rescaling Module (PRM) \cite{sajid2020plug} is to rescale the input patch ($IP$) based on its crowd-density level. As defined in \cite{sajid2020plug}, we first require the 4-way crowd-density classification (No-Crowd (NCP), low-Crowd (LCP), Medium-Crowd (MCP), High-Crowd Patch (HCP)) for input ($IP$) before using the PRM module. Thus, we branch-out the output channels from the first RB module of (Branch-1, Phase-2). These channels then proceed through the crowd-density Classification Head ($CH$) that performs the required 4-way classification (NCP, LCP, MCP, HCP). The ground-truth class-label ($CL_{IP(gt)}$) for the input patch $IP$ have been defined as follows \cite{sajid2020plug} for the training and evaluation purposes:
\begin{equation}
\label{eq1}
CL_{IP(gt)} =
    \begin{cases}
      NCP & \text{$cc_{gt}$ = 0}\\
      LCP & \text{0 \textless\ $cc_{gt}$ $\leq$\ 0.05 * $cc_{max}$}\\
      MCP & \text{0.05 * $cc_{max}$ \textless\ $cc_{gt}$ $\leq$\ 0.2 * $cc_{max}$}\\
      HCP & \text{0.2 * $cc_{max}$ \textless\ $cc_{gt}$}\\
    \end{cases} 
\end{equation}
where $cc_{gt}$ and $cc_{max}$ denote the actual people count in that patch and maximum possible crowd number in any $256\times 256$ patch in the given dataset respectively. Patch with zero ground-truth crowd count is naturally categorized with the $NCP$ label. Crowd patches with at most $5\%$ count of $cc_{max}$ are labeled as $LCP$ class. Similarly, for the patch ($IP$) with ground-truth crowd count in between $5\%$ to $20\%$ of the maximum value (with $20\%$ inclusive) is considered as the  $MCP$ category. Whereas, containing more than $20\%$ crowd count of $cc_{max}$ value makes the input patch fall into the $HCP$ class label. Depending on the designated class label (${Class}_{IP}$), the PRM rescales the input patch ($IP$) accordingly as given in \cite{sajid2020plug}. Consequently, it generates one or more new $256 \times 256$ size patches. For (${Class}_{IP} = NCP,LCP,MCP$), the PRM generates ($1,1,1$) new patches respectively. In case of (${Class}_{IP}=HCP$), the PRM divides the input patch ($IP$) into four new $128 \times 128$ patches, then upscales each by $2 \times$ to output the final $256\times 256$ size patches \cite{sajid2020plug}. The PRM generated patches then separately go through the concatenation module (CMod) to generate the initial feature maps. Eventually, these features go through the concatenation process with the (Branch-1, Phase-2) second RB-block output, followed by the Bottleneck layer ($BL$) to adjust the number of channels before proceeding further. 

The $CH$ configuration is shown in Table \ref{table:table1}. It utilizes the softmax based 4-way classification activation with the cross-entropy loss, given as follows:

\begin{equation}
\label{eq3}
Loss_{CH} = -\sum_{i=1}^{4} y_{i} log (\hat{y_{i}})
\end{equation}
where $y_i$ denotes the actual class (1 or 0) and $\hat{y_{i}}$ indicates the predicted class label. Similarly, the concatenation module (CMod) configuration is shown in Table \ref{table:table1}, where it consists of several deep layers to eventually yield the final ($32\times64\times64$) channels to be used next for the concatenation. It is also worth mentioning that the input patch ($IP$), classified as the NCP label during the test time, will be automatically discarded without any further processing. This is very effective especially in the case of discarding cluttered background regions in the images (e.g. tree leaves) that look very similar to the dense-crowd region.

\begin{figure*}
	\begin{minipage}[b]{1.0\textwidth}
		\begin{center}
			\centerline{\includegraphics[width=1.0\textwidth]{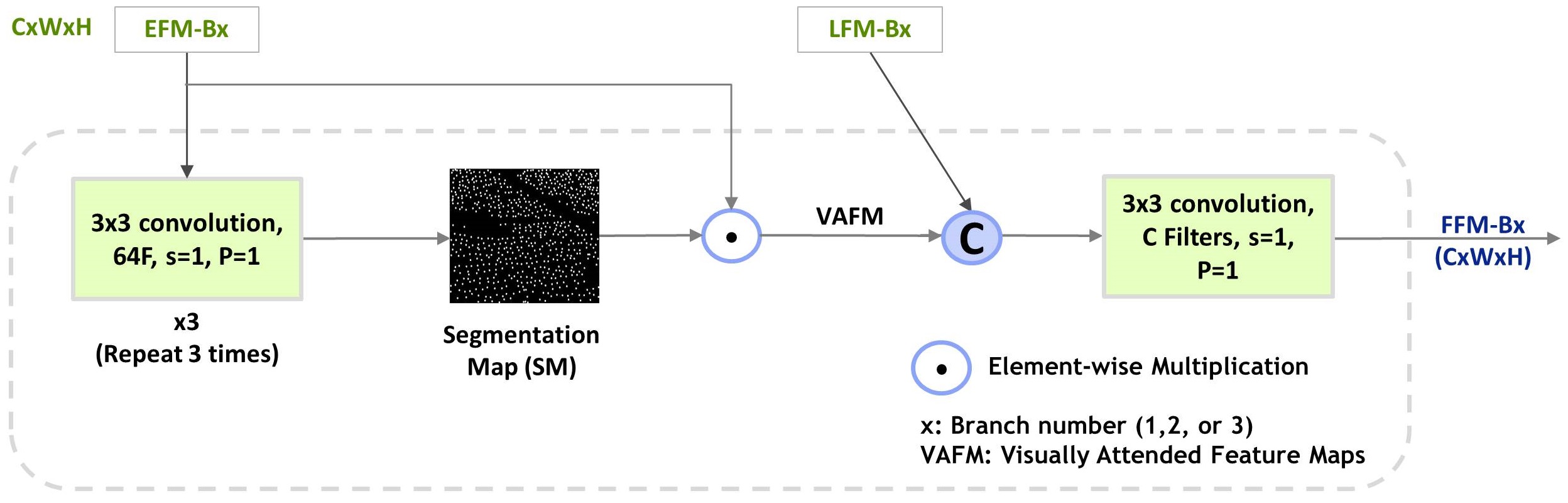}}
		\end{center}
	\end{minipage}		
	\caption{Visual Attention and Concatenation module (VACM). The early-stage feature maps (EFM-Bx) from the Branch Bx (=1,2, or 3) get visual attention and output the segmentation map (SM) that undergoes an element-wise multiplication with the original EFM-Bx channels to generate the visually-attended feature maps (VAFM). These VAFM maps concatenate with the later-stage feature maps (LFM-Bx) of the same branch before passing through the channel-adjusting deep layer to output the final feature maps (FFM-Bx) for the specific branch Bx.} 
    \label{fig:fig4}
\end{figure*}

\begin{figure}[t]
	\begin{minipage}[b]{1.0\columnwidth}
		\begin{center}
			\centerline{\includegraphics[width=1.0\columnwidth]{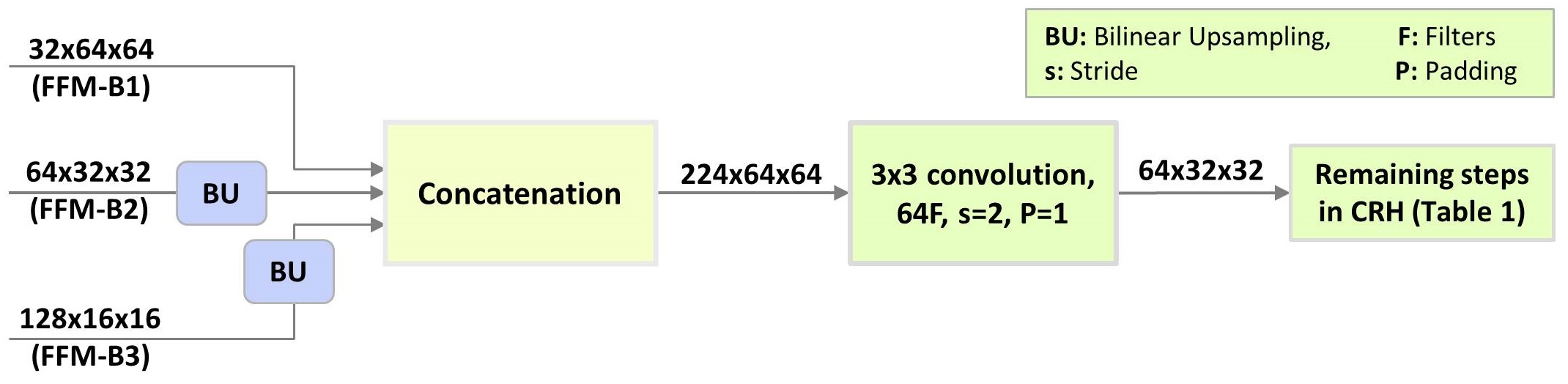}}
		\end{center}
	\end{minipage}
	\caption{\footnotesize{Concatenation-based Crowd Regression Head (CRH) concatenates the lower-level branches with the highest-level (Branch-1), followed by several deep layers to form the regression head.
	}}
    \label{fig:fig5}
\end{figure}

\subsection{Visual Attention and Concatenation Module (VACM)}
The purpose of the VACM module is to visually attend the EFM feature maps and share them with the LFM channels for better foreground vs background understanding. The VACM module visually attends the early-stage feature maps (EFM-B1, EFM-B2, EFM-B3) and concatenates them separately with their respective branch later-stage channels (LFM-B1, LFM-B2, LFM-B3). The resultant final feature maps (FFM-B1, FFM-B2, FFM-B3) then proceed through the crowd regression head (CRH) for the final crowd estimate. As shown in Fig. \ref{fig:fig4}, the input EFM-Bx block passes through three deep convolution layers to produce the attention-based segmentation map (SM $\in [0,1]_{W\times H}$). The SM undergoes element-wise multiplication with the original EFM-Bx block to produce visually attended feature maps (VAFM). These VAFMs are then concatenated with the input LFM-Bx block and channel adjusted to output final feature maps (FFM-Bx) for the branch Bx.

This early-attention mechanism helps in strengthening the early-stage feature maps of each branch towards better understanding and distinguishing the spatial foreground in comparison to the background. More importantly, this information becomes an integral part of the later-stage channels collective knowledge by concatenation-based sharing with them. Consequently, later-stage channels use this information to boost areas of interest and neglect background pixels. The SM weights are trained using the cross-entropy error ($Loss_{SM}$) between the SM and the ground-truth map. To compute the ground-truth map, we use each person's localization information already available in the benchmark datasets. This attention process significantly improves the network performance as shown in the ablation study in Sec. \ref{abl_study}.

\subsection{Crowd Regression Head ($CRH$)}
The VACM module outputs three separate sets of final feature maps (FFB-B1, FFB-B2, FFB-B3), each for the respective branch. These channels are then routed to the Crowd Regression Head (CRH). The CRH concatenates the FFB-B2 and FFB-B3 outputs with the FFB-B1 output channels using the bilinear upsampling (BU) as shown in Fig. \ref{fig:fig5}. Following that, it passes through various deep convolution and Fully Connected (FC) layers and eventually through the final 1-dimensional single neuron (1D, FC) to regress the final crowd count (${CC}_{IP}$) for the input patch ($IP$) as shown in Table \ref{table:table1}. When class label ${Class}_{IP}= HCP$, then the crowd count for the input patch ($IP$) is the sum of all four PRM generated and rescaled ($256 \times 256$ size) patches, given as follows:

\begin{equation}
\label{eq3}
\begin{split}
CC_{IP} = \sum_{i=1}^{4} cc_{p(i)} ,\ (if\  Class_{IP} = HCP)
\end{split}
\end{equation}
where $cc_{p(i)}$ with ($i=1,2,3\ and\ 4$) denotes the four rescaled patches created by the PRM for the input patch ($IP$) being classified with the $HCP$ label. The regressor utilizes Mean Square Error (MSE) as its training loss function, defined as follows:

\begin{equation}
\label{eq3}
Loss_{Regressor} = \frac{1}{T} \sum_{s=1}^{T} (F(x_s,\Theta)-y_s)^{2}
\end{equation}
where $T$ represents the total training samples per batch, $y_s$ indicates the actual crowd number for the input patch $x_s$, and $F(.)$ represents the mapping or transformation function with the learnable weights parameters $\Theta$ that learns to regress the crowd number for the input patch. Finally, the total network loss ($Loss_{total}$) is the sum of 4-way classification, segmentation map (SM), and the crowd regression losses:

\begin{equation}
\label{eq4}
Loss_{total} = Loss_{Regressor}\ +\ Loss_{CH} +\ Loss_{SM} 
\end{equation}

\section{Evaluation and Training Details}
We utilize the commonly used crowd counting metrics for numerical evaluation: Mean Absolute Error ($MAE$) and Root Mean Squared Error ($RMSE$), as defined below:
\begin{equation}
\label{eq3}
MAE = \frac{1}{TI} \sum_{h=1}^{TI} |CC_{h}-\hat{CC_{h}}|
\end{equation}

\begin{equation}
\label{eq5}
RMSE =\sqrt[]{ \frac{1}{TI} \sum_{h=1}^{TI} (CC_{h}-\hat{CC_{h}})^{2}}
\end{equation}
where $TI$ indicates the total number of test images, and $CC_{h}\ and\ \hat{CC_{h}}$ denote the true and the predicted crowd counts respectively for the test image $h$.

\textbf{Network Training Details.} To train the network, we randomly take out $75,000$ patches of $128 \times 128$, $256 \times 256$, and $512 \times 512$ sizes from the predefined training images. The resultant patches with mixed crowd numbers are resized to $256 \times 256$ size as per required. We double the training patches number by performing the horizontal-flip based data augmentation on each patch. The proposed network has been trained for 120 epochs with a batch size of 16. Stochastic gradient descent (SGD) has been used as the optimizer with a weight decay of 0.0001 and the Nesterov Momentum equal to 0.9. We also employed the multi-step learning rate ($\alpha$) that starts with the value of 0.001 and decreases to 1/2 after every 30 epochs. Also, $10\%$ training data has been separated for the network validation purpose during the model training.

\section{Experiments}
\label{expi}
This section presents both numerical and visual results based on different experiments conducted on four extensively used crowd counting benchmarks: UCF-QNRF \cite{idrees2018composition}, ShanghaiTech \cite{zhang2016single}, AHU-Crowd \cite{hu2016dense}, and the WorldExpo'10 \cite{zhang2015cross} dataset. These datasets are totally different from each other as they vary remarkably in terms of image resolution, average people count per image, maximum/minimum people per image, background regions, total images, and different lighting conditions. First, we compare our scheme numerically with the state-of-the-art models on these benchmarks. During these quantitative experiments, we also compare with the original HRNet \cite{sun2019deep,wang2020deep} model for its two independent versions: HRNet-DM and HRNet-DR. The HRNet-DM network outputs a density-map (DM) by first concatenating all branches outputs at Branch-1, followed by up-sampling it to the size of $256\times 256$. This approach is similar to the original HRNet for semantic segmentation. The final crowd-count for the input patch ($IP$) will then be the sum of all DM pixel values. Whereas, the HRNet-DR version directly regresses the final single-value crowd count, and can be considered similar to the proposed network without the VACM and PRM mechanisms, but having more residual blocks and parameters as given in the original HRNet \cite{sun2019deep,wang2020deep}. Similarly, we also report results for the proposed scheme base network (HRNet-base) without the PRM and VACM modules. Next, we present the ablation experiments and the classification head $CH$ performance analysis. Followed by the cross-dataset evaluation. Finally, the visual analysis has been presented to discuss qualitative performance.

\begin{table}[htbp]\small
	\begin{center}
	\begin{tabular}{|c|c|c|c|c|c|c|}
    \hline
 & \multicolumn{2}{c|}{ShanghaiTech-A} & \multicolumn{2}{c|}{ShanghaiTech-B} & \multicolumn{2}{c|}{UCF-QNRF}\\ \hline
Method & MAE  & RMSE & MAE  & RMSE & MAE  & RMSE\\ \hline
MCNN \cite{zhang2016single}  & 110.2  & 173.2 & 26.4  & 41.3 & 277   & 426.0   \\ \hline
CMTL \cite{cascadedmtl} & 101.3  & 152.4  & 20.0 & 31.1 & 252   & 514.0   \\ \hline
Switch-CNN \cite{sam2017switching} & 90.4 & 135.0   & 21.6 & 33.4 & 228   & 445.0   \\ \hline
SaCNN \cite{zhang2018crowd} & 86.8 & 139.2   & 16.2 & 25.8 & -   & -   \\ \hline
IG-CNN \cite{babu2018divide} & 72.5 & 118.2    & 13.6 & 21.1 & -   & -  \\ \hline
ACSCP \cite{shen2018crowd}   & 75.7  & 102.7   & 17.2 & 27.4 & -   & - \\ \hline 
CSRNet \cite{li2018csrnet}   & 68.2  & 115.0 & 10.6 & 16.0 &   -   & - \\ \hline 
CL\cite{idrees2018composition} & - & -  & -  & -  & 132 & 191.0    \\ \hline
CFF \cite{shi2019counting} & 65.2 & 109.4   & 7.2 & 12.2 & 93.8   & 146.5 \\ \hline
RRSP \cite{wan2019residual} & 63.1 & 96.2   & 8.7 & 13.6  & -   & -  \\ \hline 
CAN \cite{liu2019context} & 62.3 & 100.0   & 7.8 & 12.2 & 107   & 183.0   \\ \hline
L2SM \cite{xu2019learn} & 64.2 & 98.4   & 7.2 & 11.1 & 104.7   & 173.6 \\ \hline
BL \cite{ma2019bayesian} & 62.8 & 101.8   & 7.7 & 12.7 & 88.7 & 154.8    \\ \hline
RRP \cite{chen2020relevant} & 63.2 & 105.7   & 9.4 & 13.9 & 93 & 156.0  \\ \hline
HA-CCN \cite{sindagi2019ha} & 62.9 & 94.9   & 8.1 & 13.4 & 118.1 & 180.4    \\ \hline
ADSCNet \cite{bai2020adaptive} & \textbf{55.4} & 97.7 & \textbf{6.4} & 11.3 &   \textbf{71.3} & 132.5    \\ \hline
RPNet \cite{yang2020reverse} & 61.2 & 96.9 & 8.1 & 11.6 &   - & -    \\ \hline
ZoomCount \cite{sajid2020zoomcount} & 66.6 & 94.5 & - & - & 128 & 201.0 \\ \hline
PRM-based\cite{sajid2020plug} & 67.8 & 86.2 & 8.6 & 11.0 & 94.5 & 141.9 \\ \hline
HRNet-DM \cite{sun2019deep} & 90.5 & 111.9 & 18.7 & 29.2 &   173 & 285.4    \\ \hline
HRNet-DR \cite{sun2019deep} & 88.8 & 110.3 & 17.6 & 27.6 &   171 & 277.2    \\ \hline 
HRNet-Base & 76.3 & 110.0 & 14.2 & 21.8 & 122.5 & 219.0    \\ \hline \hline
\textbf{Ours}  & 56.1 & \textbf{79.8} & 6.6 & \textbf{9.8} & \textbf{71.3} & \textbf{120.7}   \\ \hline
Ours w/o PRM & 72.1 & 109.8 & 12.1 & 20.9 & 98.1 & 137.5    \\ \hline 

	\end{tabular}
	\end{center}
	   \caption{\footnotesize Numerical experiments on the UCF-QNRF \cite{idrees2018composition} and the ShanghaiTech \cite{zhang2016single} benchmarks. Our proposed scheme outperforms the state-of-the-art models (including the original PRM-based) under the RMSE standard criterion, while indicating closer or equal to the best results for the MAE evaluation metric. The proposed network errors increase in all cases without the PRM module usage (last row).}
	\label{table:table2}
\end{table}

\subsection{UCF-QNRF Dataset Numerical Evaluation}
The UCF-QNRF \cite{idrees2018composition} dataset contains a total of $1,535$ images with a pre-established training/testing division of 1201/334 respectively. The images contain a wide range of crowd-density and vary greatly in image resolution and background setting. The total people annotations in the dataset equal to $1,251,642$, while the images resolution varies between ($300\times377$) and ($6666\times9999$). We compare the proposed approach with the state-of-the-art models as reported in Table \ref{table:table2}. The evaluation demonstrates that our model performs the best in comparison to the state-of-the-art for the RMSE evaluation criterion with $8.9\%$ performance boost (from $132.5$ to $120.7$) amid performing equally best for the MAE metric. It may also be noted that the proposed model performs better for both metrics as compared to the original PRM based scheme (CC-2P) \cite{sajid2020plug}. Without the PRM module (last row), the proposed scheme performance degrades hugely by $27.3\%$ and $12.2\%$ in terms of the MAE and RMSE, respectively. This result demonstrates the PRM significance in the proposed model. Moreover, the HRNet-DM and HRNet-DR models prove insufficient as shown in Table \ref{table:table2}.

\subsection{ShanghaiTech Dataset Numerical Evaluation}
The ShanghaiTech \cite{zhang2016single} dataset is pre-divided into two independent parts. Part-A comprises of a total $482$ images with diverse and dense crowd range, image resolution, and varying lighting conditions. These images are already divided into $300$ training and $182$ testing images. The average image resolution is $589\times868$ with a total of $241,677$ human annotations. Total people per image is approximately $501$ on average. Part-B contains $716$ images ($400/316$ train/test split respectively) with relatively sparse crowd range and a total of $88,488$ people annotations. The mean image resolution for this part is $768\times 1024$ with an average of around $124$ people per image. We analyze the proposed scheme on this benchmark and also compare it with the state-of-the-art (including the original PRM-based \cite{sajid2020plug}) models. The results are shown in Table \ref{table:table2}, from which we can see that our model yields the best results for the RMSE criterion with an improvement of $7.4\%$ (from $86.2$ to $79.8$) and $10.9\%$ (from $11.0$ to $9.8$) on both benchmark parts respectively, and also performs reasonably well in terms of the MAE evaluation metric.
For both dataset parts, our model subjects to a huge error increase (MAE: $22.2\%$, RMSE: $27.3\%$ for Part-A, and MAE: $45.4\%$, RMSE: $53.1\%$ for Part-B) without the PRM module deployment.

\begin{table}[htbp]\small
	\begin{center}
	\begin{tabular}{|c|c|c|c|}
    \hline
Method & MAE & RMSE\\ \hline
Haar Wavelet \cite{oren1997pedestrian} & 409.0   & -    \\ \hline
DPM \cite{felzenszwalb2008discriminatively} & 395.4  & -    \\ \hline
BOW–SVM \cite{csurka2004visual} & 218.8  & -    \\ \hline
Ridge Regression \cite{chen2012feature} & 207.4  & -    \\ \hline
Hu et al. \cite{hu2016dense} & 137  & -    \\ \hline
DSRM \cite{yao2017deep} &  81  &  129    \\ \hline
ZoomCount \cite{sajid2020zoomcount} & 74.9 & 111 \\ \hline
CC-2P (PRM-based)\cite{sajid2020plug} & 66.6 & 101.9\\ \hline
HRNet-DM \cite{sun2019deep} & 80.8 & 125.8 \\ \hline
HRNet-DR \cite{sun2019deep} & 80.0 & 121.1 \\ \hline  

HRNet-Base & 78.9 & 119.3 \\ \hline  \hline
\textbf{Ours} & \textbf{57.5} & \textbf{89.0} \\ \hline
Ours w/o PRM & 76.1 & 115.3 \\ \hline

	\end{tabular}
	\end{center}

   \caption{\footnotesize AHU-Crowd \cite{hu2016dense} benchmark dataset based experiments indicate that the proposed scheme appears to be the best for both the evaluation metrics in contrast to the state-of-the-arts including the original PRM-based model. PRM-less version of our network subjects to a huge error increase in all settings as given in last row.}

	\label{table:table3}
\end{table}

\subsection{Numerical Experiments on AHU-Crowd Dataset}
AHU-Crowd dataset \cite{hu2016dense} poses a great challenge with totally different statistics as compared to the ShanghaiTech and the UCF-QNRF datasets. It only contains $107$ images with $58$ to $2,201$ ground-truth people count per image. The dataset also contains a total of $45,807$ people annotations. Based on the standard literature practice, we carried out the 5-fold cross-validation for the (MAE, RMSE) based numerical evaluation. In each fold, $96$ images were selected for training, and the remaining $11$ images for the testing purpose. We report the numerical evaluation and comparison results in Table \ref{table:table3}. These findings indicate that the proposed method outperforms other state-of-the-art methods (including the original PRM based scheme) with significant improvement of $13.7\%$ (from $66.6$ to $57.5$), $12.6\%$ (from $101.9$ to $89.0$) for the MAE and RMSE metrics respectively. Moreover, our model without the PRM module, HRNet-DM, and HRNet-DR networks subject to a huge error increase in each case as shown in Table \ref{table:table3}.

\setlength{\tabcolsep}{2.0pt}
\begin{table}[t]\small
	\begin{center}
	\begin{tabular}{|c|c|c|c|c|c|c|}
    \hline
 & S1 & S2 & S3 & S4 & S5 & Average \\ \hline
Zhang et al. \cite{zhang2015cross} & 9.8 & 14.1 & 14.3 & 22.2 & 3.7 & 12.9   \\ \hline
MCNN \cite{zhang2016single} & 3.4 & 20.6 & 12.9 & 13.0 & 8.1 & 11.6   \\ \hline
Switch-CNN \cite{sam2017switching} & 4.4 & 15.7 & 10.0 & 11.0 & 5.9 & 9.4   \\ \hline
CP-CNN \cite{sindagi2017generating} & 2.9 & 14.7 & 10.5 & 10.4 & 5.8 & 8.9  \\ \hline 
IG-CNN \cite{babu2018divide} & 2.6 & 16.1 & 10.15 & 20.2 & 7.6 & 11.3 \\ \hline
IC-CNN \cite{ranjan2018iterative} & 17.0 & 12.3 & 9.2 & 8.1 & 4.7 & 10.3 \\ \hline
CAN \cite{liu2019context}  & 2.4 & 9.4 & 8.8  & 11.2 & 4.0 & 7.2  \\ \hline 
ZoomCount \cite{sajid2020zoomcount} & 2.1 & 15.3 & 9.0  & 10.3 & 4.5 & 8.3  \\ \hline 
M-SFANet \cite{thanasutives2020encoder} & 1.88 & 13.24  & 10.07 & 7.5 & 3.87 & 7.32 \\ \hline
DSSINet \cite{liu2019crowd} & 1.57 & 9.5 & 9.46  & 10.35 & 2.49 & 6.67 \\ \hline 
ASNet \cite{jiang2020attention} & 2.22 & 10.11 & 8.89  & 7.14 & 4.84 & 6.64  \\ \hline 
PRM-based\cite{sajid2020plug} & 1.8 & 10.7 & 9.2  & 8.8 & 4.3 & 6.96  \\ \hline 
HRNet-DM \cite{sun2019deep} & 3.5 & 15.2 & 11.3 & 10.3 & 5.2 & 9.10 \\ \hline
HRNet-DR \cite{sun2019deep} & 3.5 & 14.9 & 10.8 & 10.1 & 4.9 & 8.84 \\ \hline 
HRNet-Base & 2.95 & 13.1 & 10.2 & 9.6 & 4.8 & 8.13 \\ \hline \hline
\textbf{Ours} & \textbf{1.53} & \textbf{9.04} & \textbf{8.03}  & \textbf{7.10} & \textbf{2.22} & \textbf{5.58}  \\ \hline
Ours w/o PRM & 2.19 & 10.98 & 9.75 & 9.11 & 5.08 & 7.42 \\ \hline
	\end{tabular}
	\end{center} 
	
	\caption{\footnotesize Quantitative evaluation and comparison on the WorldExpo'10 dataset \cite{zhang2015cross} demonstrate that the proposed scheme appears as the best by yields the least MAE error on all five test scenes (S1-S5) as well as their average value. Our PRM-less network (last row) subjects to significant performance degradation, thus, emphasizing its effectiveness.}
	\label{table:table4}
\end{table}

\subsection{WorldExpo’10 Benchmark Numerical Experiments}
The WorldExpo'10 \cite{zhang2015cross} dataset contains $3,980$ annotated frames, collected from $1,132$ video sequences of $108$ different scenes. Out of these frames, $3,380$ images from $103$ different scenes have been used for the training purpose. The remaining $600$ frames from $5$ different scenes work as the test split. As per the standard practice, we only utilize the pre-defined Region of Interest (RoI) as given in both training and testing images. Also, we evaluate and compare with the state-of-the-art on five test scenes using the MAE evaluation metric. As shown in Table \ref{table:table4}, the proposed approach yields the best performance on all scene images with the $16.0\%$ (from $6.64$ to $5.58$) average MAE error decrease.
The model subjects to $24.8\%$ average MAE increase (from $5.58$ to $7.42$) without the PRM module deployment, thus, signifying its importance in our network.

\subsection{Ablation Experiments Study}
\label{abl_study}
In this section, we present six ablation studies on the ShanghaiTech dataset \cite{zhang2016single} to investigate the effect of different components of the proposed scheme.

\begin{table}[htbp]\small
	\begin{center}
	\begin{tabular}{|c|c|c|c|c|}
    \hline
 \multicolumn{5}{|c|}{Total Branches Choice}\\ \hline
 \multicolumn{3}{|c|}{Total Branches} & MAE  & RMSE\\ \hline
 \multicolumn{3}{|c|}{1} & 91.6  & 131.4    \\ \hline
 \multicolumn{3}{|c|}{2} & 73.6  & 109.7    \\ \hline
 \multicolumn{3}{|c|}{\textbf{3 (our default choice)}} & \textbf{56.1}  & \textbf{79.8}    \\ \hline
 \multicolumn{3}{|c|}{4} & 69.6  & 97.5    \\ \hline \hline
 \multicolumn{5}{|c|}{Residual Units (RUs) per RB Block Quantity Effect}\\ \hline
 \multicolumn{3}{|c|}{RU units per RB block: 2} & 77.9 & 101.1   \\ \hline
 \multicolumn{3}{|c|}{3} & 71.5 & 98.3   \\ \hline
 \multicolumn{3}{|c|}{\textbf{4 (our default choice)}} & \textbf{56.1}  & \textbf{79.8}    \\ \hline
 \multicolumn{3}{|c|}{5} & 71.7 & 97.2  \\ \hline
 \multicolumn{3}{|c|}{6} & 73.6 & 102.8   \\ \hline \hline
 \multicolumn{5}{|c|}{2-Layered vs 3-Layered Residual Unit Choice}\\ \hline
 \multicolumn{3}{|c|}{2-Layered} & 70.6  & 96.9    \\ \hline
 \multicolumn{3}{|c|}{\textbf{3-Layered (our default choice)}} & \textbf{56.1}  & \textbf{79.8}    \\ \hline \hline
 \multicolumn{5}{|c|}{Branching-out Positioning in the Network}\\ \hline
 \multicolumn{3}{|c|}{From RB (of Phase1)}  & 71.0 & 97.9   \\ \hline
 \multicolumn{3}{|c|}{\textbf{1st RB (of Branch-1, Phase2)}} & \textbf{56.1} & \textbf{79.8}   \\ \hline
 \multicolumn{3}{|c|}{2nd RB (of Branch-1, Phase2)} & 69.1 & 94.8   \\ \hline
 \multicolumn{3}{|c|}{1st RB (of Branch-1, Phase3)} & 70.9 & 96.1   \\ \hline
\multicolumn{5}{|c|}{Visual Attention (VACM) Effect}\\ \hline
 \multicolumn{3}{|c|}{w/o VACM} & 63.9  & 82.4    \\ \hline
 \multicolumn{3}{|c|}{\textbf{w VACM (our default choice)}} & \textbf{56.1}  & \textbf{79.8}    \\ \hline \hline
 
 \multicolumn{5}{|c|}{EFM-Bx and LFM-Bx choices for SM, VACM Generation and Concatenation}\\ \hline
 \multicolumn{2}{|c|}{Features used for SM \& VAFM Generation} & For Concatenation & MAE  & RMSE    \\ \hline
 \multicolumn{2}{|c|}{\textbf{EFM-Bx (our default)}} & \textbf{LFM-Bx (our default)} & \textbf{56.1}  & \textbf{79.8}    \\ \hline
 \multicolumn{2}{|c|}{LFM-Bx} & LFM-Bx & 59.4  & 81.0   \\ \hline
 \multicolumn{2}{|c|}{LFM-Bx} & EFM-Bx & 58.8  & 80.5   \\ \hline
	\end{tabular}
	\end{center}
	
	\caption{\footnotesize Six different sets of ablation experiments validate our selection of the few vital hyper-parameters for the proposed network.}
	\label{table:table5}
\end{table}

\begin{figure}[t]
	\begin{minipage}[b]{1.0\columnwidth}
		\begin{center}
			\centerline{\includegraphics[width=1.0\columnwidth]{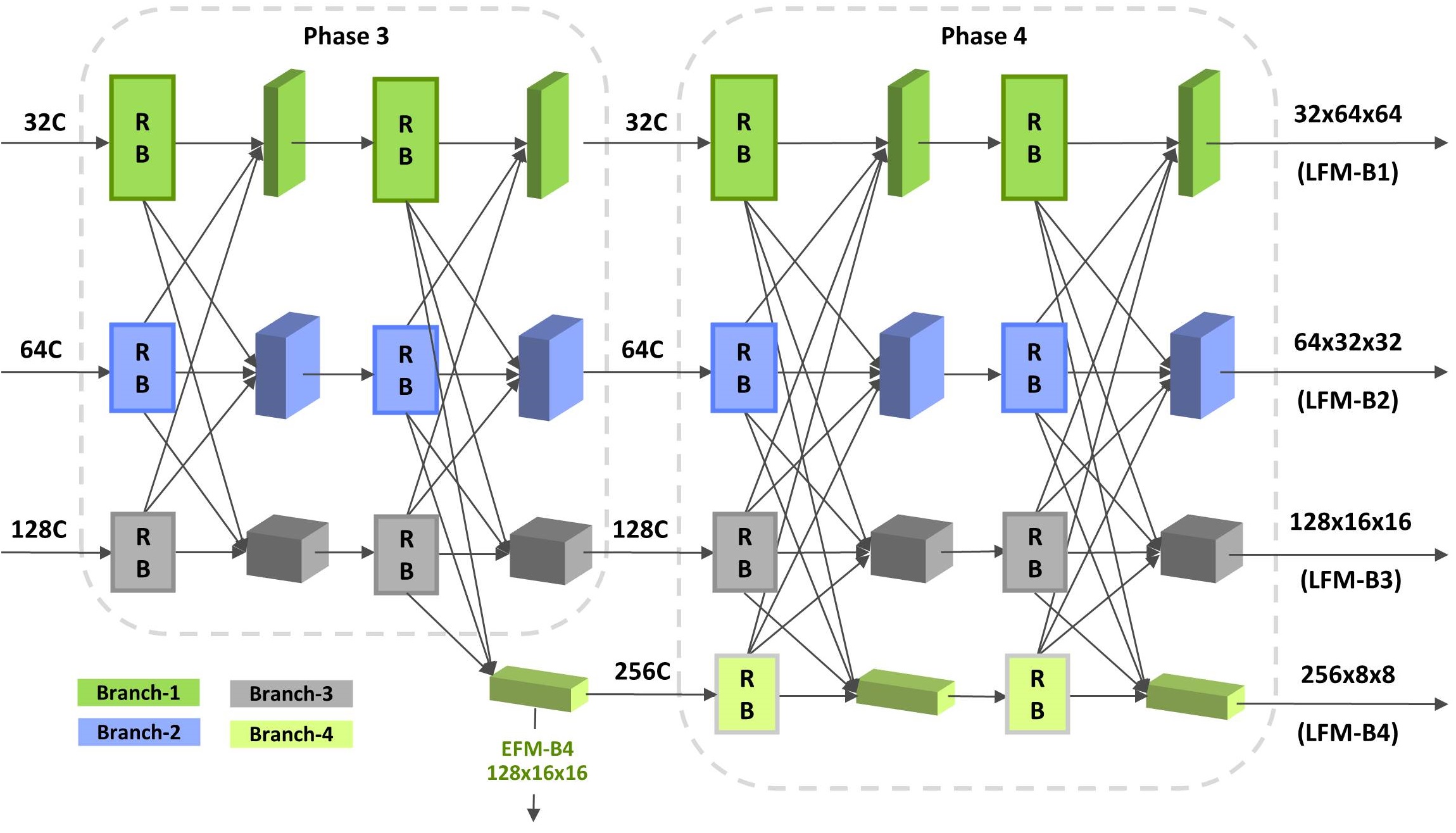}}
		\end{center}
	\end{minipage}
	
	\caption{\footnotesize{Four-branch based version of the proposed network for the branch quantity ablation study. In this setting, a new branch (Branch-4) has been added into the proposed (three-branch) model that naturally results in a new phase (Phase-4) as well. Also, one additional EFM (EFM-B4) block emerges from Branch-4 that passes through VACM along with its respective LFM block (LFM-B4). The outputs from Phase-4 now serve as the LFM feature maps for the remaining crowd estimation process.
	}}
    \label{fig:fig6}
\end{figure}

\textit{Effect of the total number of Columns/Branches.} First ablation study discusses the consequences related to the quantity of multi-resolution branches (or columns) being used in the proposed network. We explore this critical hyper-parameter by experimenting separately with different quantities of such branches. The study results are shown in Table \ref{table:table5}. Using only one branch ($Branch-1$) produces the worst results, since it only contains a single column and thus lacks any fusion or information-sharing. Two-Branch ($Branch-1, Branch-2$) network does not contain the $Branch-3$ column, but performs better than one-branch based model. The proposed network, with three multi-resolution branches, gives the best performance in comparison to the above configurations as well as the model with four multi-resolution branches as indicated in the same table. Due to this ablation study outcome, the proposed model has been designed with three multi-resolution branches. In the four-branch network experiment, we deployed an additional branch ($Branch-4$) with $2\times$ down-scaled resolution ($8\times8$) and double the channels ($256$) than the $Branch-3$ as shown in Fig. \ref{fig:fig6}. The four-branch based model naturally contains an additional phase ($Phase4$) to cover the fusion process for $Branch-4$ with other branches. Naturally, it also contains one more early-stage block (EFM-B4) emerging from Branch-4. The EFM-B4 is routed to the VACM module along with its respective later-stage block (LFM-B4). Additionally, the $Phase4$ outputs now serve as the LFM blocks.

\textit{Effect of the number of residual units in RB blocks.} Here, we investigate the effect of using four 3-layered residual units (RU) per RB block as compared to deploying other potential quantities ($2,3,5,$ or $6$ RU units per RB block). As shown in the ablation experiments results in Table \ref{table:table5}, 4 RU units per RB block yield the best results with the lowest MAE and RMSE errors. Thus, it acts as our default and preferred choice in the proposed network.

\textit{Effect of using 3-layered vs 2-layered residual unit.} We have two major choices for the Residual Unit: 2-layered or 3-layered deep residual unit as given in \cite{he2016deep}. Results for both choices are shown in Table \ref{table:table5}. It is evident that the 3-layered RU performs much better than the 2-layered residual unit.

\textit{Effect of branching-out location in the network.} As shown in the proposed network in Fig. \ref{fig:fig2}, we branch-out the output features of the first RB block in the (Branch-1, Phase-2) to feed into the classification head. Here, we investigate the effect of the location of this branching-out by re-positioning it to other Branch-1 RB blocks output. As shown in Table \ref{table:table5}, our default choice of branching-out from 1st RB of (Branch-1, Phase-2) gives the lowest MAE and RMSE error. Additionally, in each ablation experiment setting, the concatenation back into the network happens with the RB block of the Branch-1 that is next and subsequent to the RB block responsible for the branching-out.

\textit{VACM Module Effect.} Visual attention on the early-stage channels helps in enriching the later-stage feature maps. Consequently, the VACM module should improve the overall network performance. As shown in Table \ref{table:table5}, the visual attention process boosts the network effectiveness by ($12.2\%$, $3.2\%$) in terms of (MAE, RMSE) respectively.

\textit{EFM-Bx and LFM-Bx usage choices effect in the VACM module.} Here, we investigate the effect of deploying the EFM-Bx and LFM-Bx features in different combinations for SM and VAFM features generation and the subsequent concatenation process in the VACM module. The results are shown in Table \ref{table:table5}, where the EFM-Bx and LFM-Bx features have been used in different settings for SM and VAFM generation as well as the concatenation process. As shown, the proposed combination of deploying EFM-Bx features for SM and VAFM generation, followed by concatenating them into the LFM-Bx features empirically proves to be the most effective choice.

\begin{figure}[t]
	\begin{minipage}[b]{1.0\columnwidth}
		\begin{center}
			\centerline{\includegraphics[width=1.0\columnwidth]{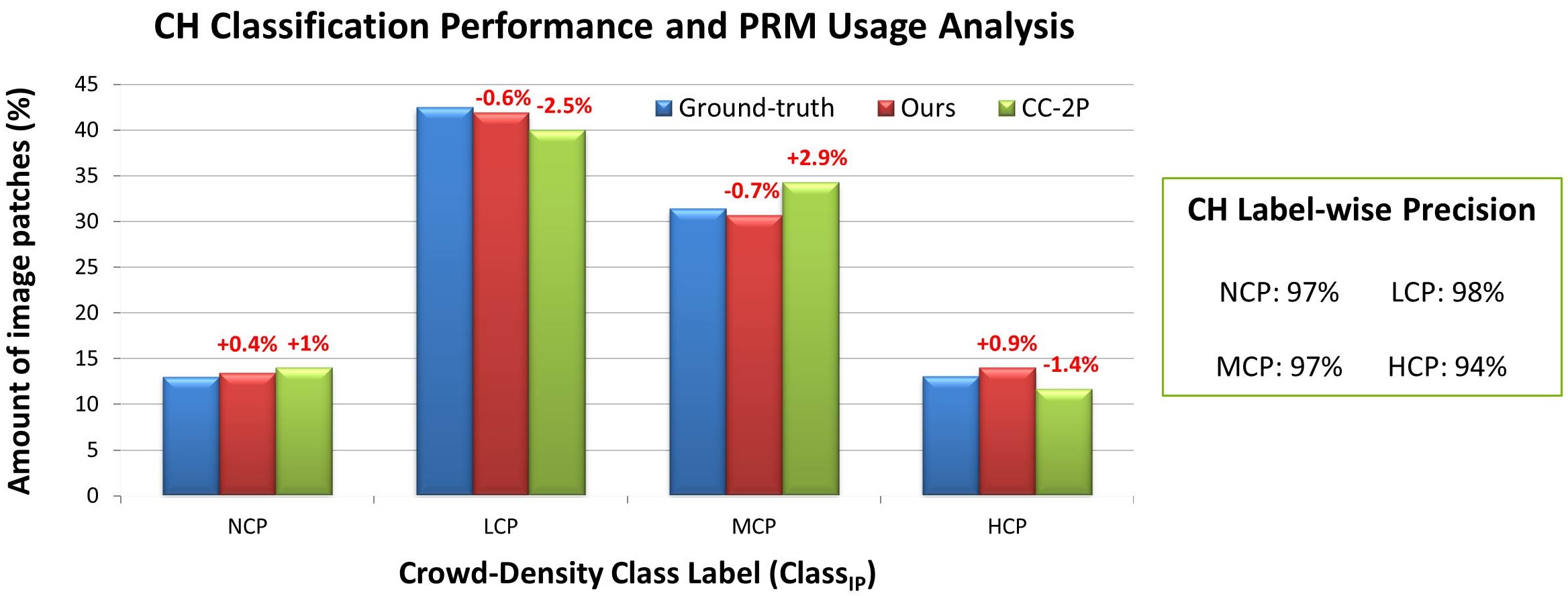}}
		\end{center}
	\end{minipage}
	
	\caption{\footnotesize{The classification head ($CH$) effectiveness and PRM label-wise usage analysis on the ShanghaiTech-A dataset test image patches. The left graph indicates that the $CH$ head performs better than the state-of-the-art CC-2P \cite{sajid2020plug} model classifier on all four labels (NCP,LCP,MCP,HCP). It also demonstrates that a huge percentage of patches ($55.9\%$) are designated with either $LCP$ or $HCP$, thus, emphasizing frequent use of the subsequent PRM rescaling operation. The $CH$ also performs reasonably well in terms of label-wise precision (right block).
	}}
    \label{fig:fig7}
\end{figure}

\begin{table}[htbp]\small 
	\begin{center}
	\begin{tabular}{|c|c|c|}
    \hline
Method & MAE  & RMSE\\ \hline
Switch-CNN \cite{sam2017switching} & 301  & 457   \\ \hline
Cascaded-mtl \cite{cascadedmtl} & 308  & 478  \\ \hline
CC-2P (PRM based) \cite{sajid2020plug} & 219 & 305    \\ \hline 
\textbf{Ours} & \textbf{197} & \textbf{271}    \\ \hline
	\end{tabular}
	\end{center}
	
	\caption{\footnotesize Based on the comparison with the state-of-the-art methods during the cross-dataset evaluation, our approach outperforms them under both evaluation criteria.}
	\label{table:table6}
\end{table}

\subsection{Cross-Dataset Evaluation}
To further assess the proposed model, we conducted the cross-dataset evaluation. ShanghaiTech \cite{zhang2016single} benchmark has been used for all models training, while the testing has been conducted using the UCF-QNRF dataset. Table \ref{table:table6} reports the proposed approach cross-dataset performance and also compares it to other state-of-the-art schemes. It is clear from the results that the proposed model outperforms other methods including the original PRM-based scheme. These findings also indicate the better generalization potential of our scheme towards unseen images with different dynamics and crowd diversity.

\subsection{Classification Head ($CH$) performance and PRM class-wise usage analysis}
The classification head ($CH$) plays a pivotal role in the success of the subsequent PRM process. Here, we investigate and compare our classification head ($CH$) effectiveness with the state-of-the-art PRM-based CC-2P model \cite{sajid2020plug} classifier on the ShanghaiTech-A \cite{zhang2016single} benchmark test images patches. The results for all four crowd-labels ($NCP,LCP,MCP,HCP$) are shown in Fig. \ref{fig:fig7}, where we can see that our $CH$ classifier yields more accurate and closer to the ground-truth results for each class-label as compared to the CC-2P classifier. On the other hand, this analysis also gives insight into the percentage of image patches utilizing any specific PRM re-scaling operation. As shown in Fig. \ref{fig:fig7}, around $41.9\%$ and $14\%$ of test image patches have been classified as $LCP$ and $HCP$ respectively, and consequently require appropriate PRM-based re-scaling. Similarly, $\sim13\%$ of the test patches have been discarded as being classified with the $NCP$ label that may have resulted in huge crowd over-estimation otherwise. Thus, the $CH$ head proves more effective in rightly categorizing all crowd-density labels. The PRM's main re-scaling operations (Up- and Down-scaling) are required in most cases ($55.9\%$), therefore, emphasizing its great importance in the proposed scheme. The $CH$ head also demonstrates reasonable performance in terms of label-wise precision (with minimum $94\%$ precision) as shown on the right in Fig. \ref{fig:fig7}. This is partly due to better training of Phase-1 RB block with the aid of inter-branch fusion and the VACM process.

\begin{figure*}
	\begin{minipage}[b][][b]{0.245\columnwidth}
		\begin{center}
			\centerline{\includegraphics[width=1\columnwidth]{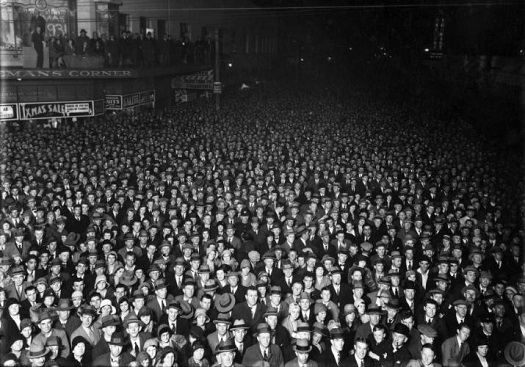}}
			\centerline{\footnotesize{GT=1443, PRM=1131}}
			\centerline{\footnotesize{Ours=1427, DME=388}}
		\end{center}
	\end{minipage}
	\begin{minipage}[b][][b]{0.245\columnwidth}
		\begin{center}
			\centerline{\includegraphics[width=1\columnwidth]{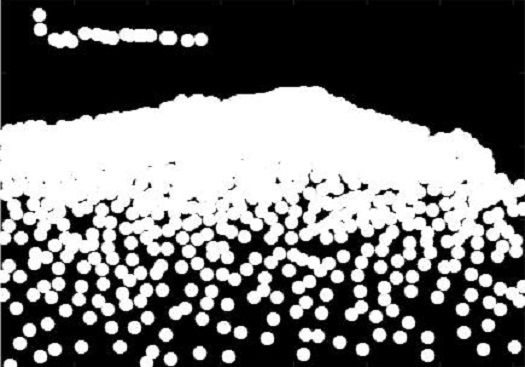}}
			\centerline{\footnotesize{}}
			\centerline{\footnotesize{}}
		\end{center}
	\end{minipage}
		\begin{minipage}[b][][b]{0.245\columnwidth}
		\begin{center}
			\centerline{\includegraphics[width=1\columnwidth]{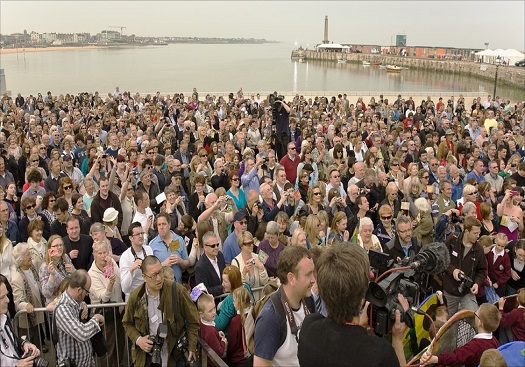}}
			\centerline{\footnotesize{GT=1103, PRM=908}}
			\centerline{\footnotesize{Ours=1089, DME=842}}
		\end{center}
	\end{minipage}
	\begin{minipage}[b][][b]{0.245\columnwidth}
		\begin{center}
			\centerline{\includegraphics[width=1\columnwidth]{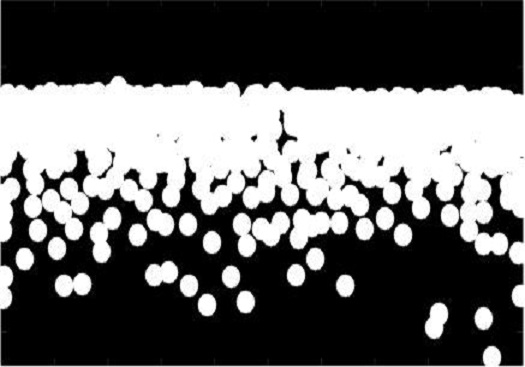}}
			\centerline{\footnotesize{}}
			\centerline{\footnotesize{}}
		\end{center}
	\end{minipage}
	\begin{minipage}[b][][b]{0.245\columnwidth}
		\begin{center}
			\centerline{\includegraphics[width=1\columnwidth]{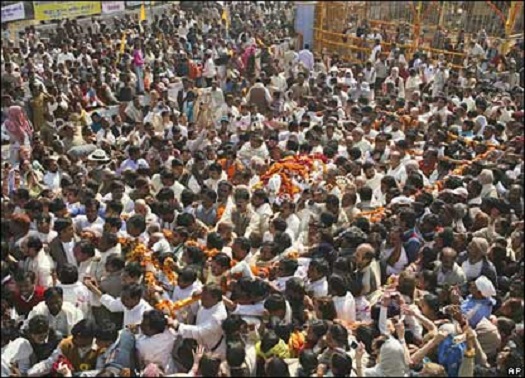}}
			\centerline{\footnotesize{GT=556, PRM=478}}
			\centerline{\footnotesize{Ours=555, DME=236}}
		\end{center}
	\end{minipage}
	\begin{minipage}[b][][b]{0.245\columnwidth}
		\begin{center}
			\centerline{\includegraphics[width=1\columnwidth]{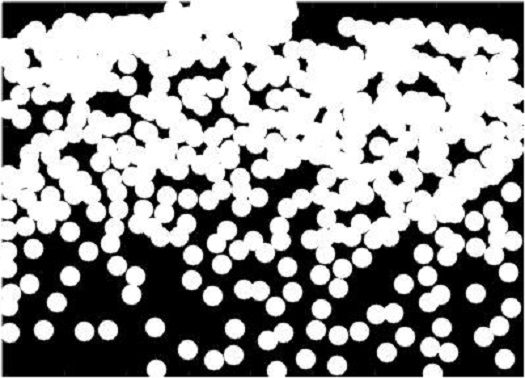}}
			\centerline{\footnotesize{}}
			\centerline{\footnotesize{}}
		\end{center}
	\end{minipage}
	\begin{minipage}[b][][b]{0.245\columnwidth}
		\begin{center}
			\centerline{\includegraphics[width=1\columnwidth]{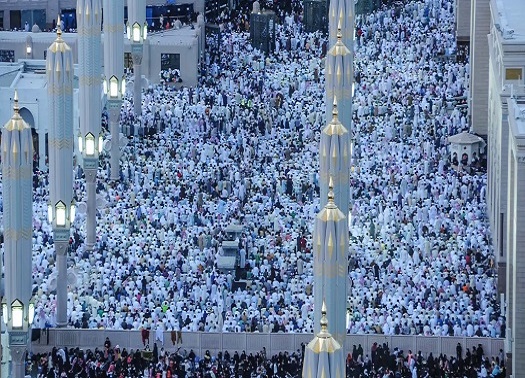}}
			\centerline{\footnotesize{GT=3653, PRM=2992}}
			\centerline{\footnotesize{Ours=3609, DME=1692}}
		\end{center}
	\end{minipage}
	\begin{minipage}[b][][b]{0.245\columnwidth}
		\begin{center}
			\centerline{\includegraphics[width=1\columnwidth]{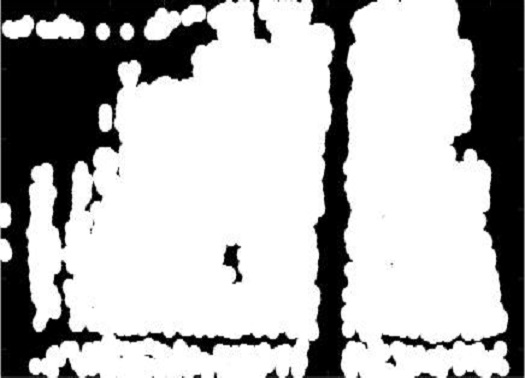}}
			\centerline{\footnotesize{}}
			\centerline{\footnotesize{}}
		\end{center}
	\end{minipage}
		\begin{minipage}[b][][b]{0.245\columnwidth}
		\begin{center}
			\centerline{\includegraphics[width=1\columnwidth]{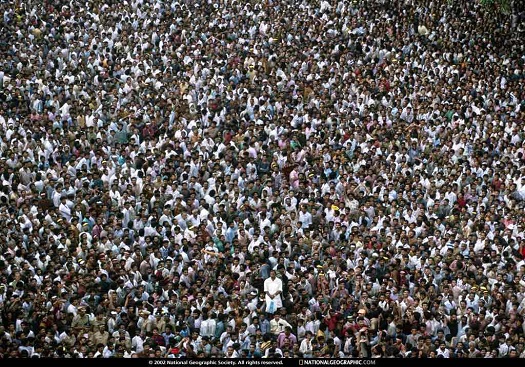}}
			\centerline{\footnotesize{GT=2472, PRM=2017}}
			\centerline{\footnotesize{Ours=2488, DME=1370}}
		\end{center}
	\end{minipage}
	\begin{minipage}[b][][b]{0.245\columnwidth}
		\begin{center}
			\centerline{\includegraphics[width=1\columnwidth]{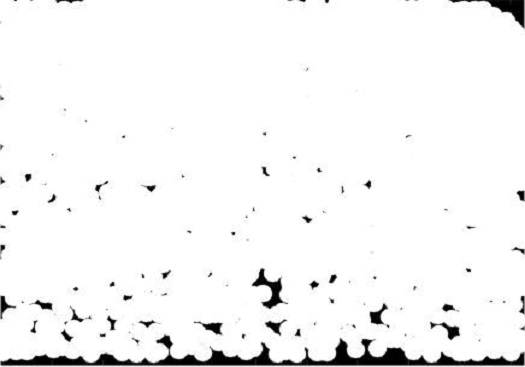}}
			\centerline{\footnotesize{}}
			\centerline{\footnotesize{}}
		\end{center}
	\end{minipage}
	\begin{minipage}[b][][b]{0.245\columnwidth}
		\begin{center}
			\centerline{\includegraphics[width=1\columnwidth]{figs/fig8f.jpg}}
			\centerline{\footnotesize{GT=207, PRM=130}}
			\centerline{\footnotesize{Ours=204, DME=109}}
		\end{center}
	\end{minipage}
	\begin{minipage}[b][][b]{0.245\columnwidth}
		\begin{center}
			\centerline{\includegraphics[width=1\columnwidth]{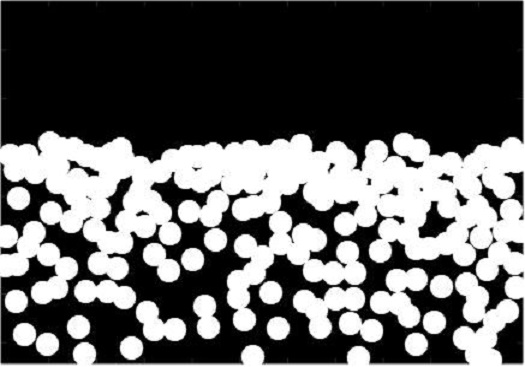}}
			\centerline{\footnotesize{}}
			\centerline{\footnotesize{}}
		\end{center}
	\end{minipage}
    \vspace{-6mm}
	\caption{\footnotesize{Visual evaluation. The proposed scheme generates better crowd counts than the state-of-the-art PRM-based \cite{sajid2020plug} and the density-map estimation (DME)  \cite{idrees2018composition} based methods when compared in terms of the ground truth (GT). For each test image, the VACM generated segmentation-map (SM) also indicates reasonable foreground information capture that is being utilized by subsequent spatial visual attention process.
	}}
	\label{fig:fig8}
\end{figure*}

\begin{table*}[t]\small
	\begin{center}
	\begin{tabular}{|c|c|c|c|c|c|c|}
    \hline
Method & $T_{total}$ (secs) & $T_{avg}$ (secs) & $Parameters$ (millions)  & MAE & RMSE \\ \hline

CSRNet \cite{li2018csrnet} & 60.2 & 0.33  & 16.3 & 68.2 & 115.0   \\ \hline
CP-CNN \cite{sindagi2017generating} & 122.8 & 0.68 & 68.4  & 73.6 & 106.4   \\ \hline
ZoomCount \cite{sajid2020zoomcount} & 85.4 & 0.47 & 14.0  & 66.6 & 94.5   \\ \hline
CC-2P \cite{sajid2020plug} & 55.4 & 0.30 & \textbf{5.1}  & 67.8 & 86.2 \\ \hline \hline
Ours & \textbf{49.7} & \textbf{0.27} & 25.6 & \textbf{56.1} & \textbf{79.8}   \\ \hline

	\end{tabular}
	\end{center} 
	\caption{\footnotesize Inference speed and total model parameters analysis on ShanghaiTech Part-A dataset.}
	\label{table:table7}
\end{table*}

\subsection{Inference Speed Analysis}
In this experiment, we perform inference speed analysis and comparison with four state-of-the-art schemes (CSRNet \cite{li2018csrnet}, CP-CNN \cite{sindagi2017generating}, ZoomCount \cite{sajid2020zoomcount}, PRM-based CC-2P \cite{sajid2020plug}) as given in Table \ref{table:table7}. $T_{total}$ and $T_{avg}$ denote total and average time taken by the models on $182$ test images of ShanghaiTech Part-A benchmark. NVIDIA Titan Xp GPU has been used in all settings. As shown in this result, the proposed scheme appears as the most efficient and effective one with the least $T_{total}$, $T_{avg}$, MAE, and RMSE values. Better efficiency is achieved due to the parallel structure of the proposed network. Moreover, in the high-density case (${Class}_{IP}=HCP$), all four  PRM-generated independent patches are also processed in parallel in the next steps, thus, increasing the efficiency. We also report the total learnable network parameters in the same table. The PRM-based CC-2P network contains the least parameters. The proposed scheme contains relatively much more parameters, but yields better efficiency and speed.

\subsection{Visual Analysis}
In this section, we show a few visual results. Six original test images are shown in Fig. \ref{fig:fig8}, where we analyze our scheme against the state-of-the-art PRM \cite{sajid2020plug} and the density-map estimation (DME) \cite{idrees2018composition} based methods. For each test image, we show the original image as well as the segmentation map (SM) being generated in the VACM module. These images contain hugely varying crowd-density and scale with fluctuating lighting conditions and background. As shown in Fig. \ref{fig:fig8}, the proposed scheme yields the best results that are closer to the ground-truth in comparison to the two competing models. For the visualization purpose, each SM is thresholded at 0.5 to make it a binary segmentation map. As it can be seen qualitatively, the early-stage feature maps generate these highly accurate maps and help in foreground information capture that is being utilized by subsequent spatial visual attention process.

\section{Conclusion}
We have proposed a new multi-resolution fusion and multi-task based crowd counting network with visual attention in this paper by further exploring and more effectively utilizing the PRM module. The proposed method relies on the PRM module and builds the collective knowledge using the feature-level fusion across the multi-resolution branches as well as visually attending the early-stage channels to boost the foreground vs background understanding of later-stage channels. This integration technique outperforms the state-of-the-art (including the original PRM based) methods as demonstrated through extensive standard numerical and visual experiments and comparisons. The proposed scheme also demonstrates better generalization ability during the cross-dataset evaluations.

\section*{Acknowledgement}

This work was partly supported in part by the Natural Sciences and Engineering Research Council of Canada (NSERC) under grant no. RGPIN-2021-04244, and the United States Department of Agriculture (USDA) under Grant no. 2019-67021-28996. We would also like to thank Xi Mo for proofreading the paper.

\bibliography{mybibfile}

\end{document}